\documentclass[conference]{IEEEtran}
\IEEEoverridecommandlockouts
\usepackage{cite}
\usepackage{amsmath,amssymb,amsfonts}
\usepackage{algorithmic}
\usepackage{graphicx}
\usepackage{textcomp}
\usepackage{xcolor}
\usepackage{booktabs}
\def\BibTeX{{\rm B\kern-.05em{\sc i\kern-.025em b}\kern-.08em
    T\kern-.1667em\lower.7ex\hbox{E}\kern-.125emX}}
\begin{document}

\title{Interpretable EEG biomarkers with bag-of-waves:\\ Spatial and temporal waveform dictionaries for low-data regimes}

\author{
\IEEEauthorblockN{Athanasios Papastathopoulos-Katsaros\textsuperscript{1,2},
Steven T. Lee\textsuperscript{1,2,3},
Lin Yao\textsuperscript{1,2,3},
Ajay Thomas\textsuperscript{1,3},\\
Junseok Park\textsuperscript{1,2},
Matthew J. McGinley\textsuperscript{2,4,5,6},
and Zhandong Liu\textsuperscript{1,2}}
\IEEEauthorblockA{
\textsuperscript{1}\textit{Department of Pediatrics, Baylor College of Medicine}, Houston, TX, USA \\
\textsuperscript{2}\textit{Jan and Dan Duncan Neurological Research Institute, Texas Children's Hospital}, Houston, TX, USA \\
\textsuperscript{3}\textit{Department of Child Neurology and Developmental Neurosciences, Texas Children's Hospital}, Houston, TX, USA \\
\textsuperscript{4}\textit{Department of Neuroscience, Baylor College of Medicine}, Houston, TX, USA \\
\textsuperscript{5}\textit{Gordon and Mary Cain Pediatric Neurology Research Foundation Laboratories, Texas Children's Hospital}, Houston, TX, USA \\
\textsuperscript{6}\textit{Department of Electrical and Computer Engineering, Rice University}, Houston, TX, USA}}

\maketitle

\begin{abstract}
Electroencephalography (EEG) is widely used to diagnose neurological conditions, but its analysis usually relies on either predefined spectral features or deep neural networks. Predefined features carry a strong bias, since they fix in advance what counts as informative, while deep neural networks and foundation models are hard to interpret and need large amounts of data and compute. We present bag-of-waves, an interpretable framework that learns a small dictionary of recurring EEG waveform templates, called atoms, using shift-invariant k-means without labels. The continuous EEG is then turned into a sequence of atom tokens, whose counts feed a simple downstream classifier or clustering step. We extend this representation in two ways: we add atom-to-atom transitions, which we call n-grams, to capture temporal structure, and we move from single-channel atoms to regional and cross-channel spatial atoms for the multichannel case. We test the method on three complementary datasets, each probing a different aspect: single-channel mouse genotype clustering with only sixteen animals (the low-data and temporal case), resting-state dementia classification (the spatial case), and the TUEV benchmark, a six-way classification of clinical EEG events (a high-data comparison against strong deep and foundation baselines). Across all three datasets, bag-of-waves achieves performance competitive with state-of-the-art deep and foundation models. Yet, it operates with a fraction of the parameter count and provides full interpretability: because every atom corresponds to an inspectable waveform, the method explicitly recovers known clinical morphologies that a neurophysiologist can directly validate. Its main advantage is that it works in the low-data regime where heavier models are a poor fit.
\end{abstract}


\section{Introduction}
Electroencephalography (EEG) records the electrical potential from the joint activity of many neurons~\cite{b_buzsaki_fields}, usually from electrodes on the scalp, while the subject is at rest or performing a task. It contains rhythmic oscillations in known frequency bands together with transient sharp waves and spikes, and these patterns change in many neurological conditions, so EEG is used extensively to diagnose them~\cite{b_guo} (and epilepsy in particular).

Three broad approaches analyze it. Visual inspection by a clinician is interpretable~\cite{b_kane} but manual and subjective. Quantitative EEG (qEEG) reduces a recording to predefined features such as band power, spectral ratios, or coherence~\cite{b_qeeg}, which are reproducible but fix in advance what counts as informative, can miss the morphology of the underlying waveforms, and are less interpretable than the patterns a clinician reads. Machine learning, supervised or unsupervised, is used mostly in research but is emerging in the clinic. Within it, two families dominate: classifiers built on the same predefined qEEG features~\cite{b_qeegrev}, and deep neural networks with large pretrained foundation models~\cite{b_biot,b_labram}, which reach high accuracy when data is abundant but are hard to interpret, costly to train, and usually trained on human recordings with standard montages such as the 10-20 system, limiting transfer to animal models, other recording setups, and patients with rare phenotypes where data is scarce.

For this reason, we present a method that fills this gap: it is interpretable, requires little data, is lightweight, and works across recording setups and clinical questions. To do this, we adapt the bag-of-words idea from text retrieval~\cite{b_tfidf}, where a document is summarized by how often each word appears. We learn a dictionary of recurring waveform patterns from the EEG, then tokenize a recording by assigning each short window to its best-matching pattern, and run the downstream model on the resulting statistics. We call each learned waveform an atom, token, or word, and a short ordered sequence of atoms an n-gram. Dictionary learning is label-free, matching is shift-invariant so a waveform is recognized regardless of where it starts, the n-grams capture temporal dynamics when the task needs them, and every atom and transition is a waveform a clinician can inspect.

We extend the single-channel bag-of-waves classifier of Cano Achuri et al.~\cite{b_bow_mouse} in two ways. First, we move from single-channel atoms to regional and cross-channel spatial atoms, so an atom can capture structure across channels. Second, we add atom-to-atom transitions, the n-grams above, so the representation can describe how waveforms follow one another over time. A further design choice is to keep the dictionary small: we deliberately restrict the method to a small set of waveforms, both for interpretability and because a small dictionary keeps the token counts dense and reproducible.

Several lines of work are related. Earlier EEG waveform dictionaries learn atoms through sparse, shift-invariant coding and matching pursuit~\cite{b_lewicki,b_brockmeier,b_mendoza}, and multivariate versions extend this to several channels at once~\cite{b_barthelemy,b_dupre}. These methods reconstruct the signal as a sum of several atoms with continuous coefficients, whereas we assign a single atom to each window and use the discrete counts, not a reconstruction, as features. They also do not model how atoms follow one another. Modeling temporal structure through such transitions is done in EEG microstate analysis~\cite{b_microstate} and in hidden Markov models over brain states~\cite{b_hmm}, but here we use discrete, inspectable waveform tokens with explicit n-gram counts rather than latent states and a learned transition matrix. In this sense a spatial atom is close to an EEG microstate, a recurring cross-channel pattern, except that atoms are learned waveforms matched shift-invariantly rather than instantaneous topographies.

Our central claim is that a small, unsupervised dictionary (with $K$ between 6 and 64 atoms), either regional or spatial, together with the transitions between its atoms, can capture enough of the signal to classify well in downstream tasks while remaining interpretable. By unsupervised we mean only the dictionary learning: the atoms are learned without labels, and a separate downstream step, supervised classification or unsupervised clustering, is then fit on top. Learning the dictionary without labels is deliberate, because in clinical settings the recordings are often available long before the annotations are, so a label-free feature extractor can be built first and reused within a dataset as labels arrive.

We test the method on three datasets, each representing a different challenge. We compare against published deep and foundation models, which serve only as a reference range, since beating them is not our goal.

Our contributions are threefold: we extend bag-of-waves to regional and cross-channel spatial atoms and to atom-to-atom transitions (n-grams); we show that a small dictionary keeps token counts dense and reproducible, which makes the transition features usable on small samples; and we show on three datasets that the label-free atoms recover known morphologies, including genotype-specific background morphology in mice, alpha slowing in Alzheimer's disease, and periodic discharges (PLED and GPED) on TUEV.

\section{Method}
The pipeline has three stages: a dictionary of waveform atoms is learned across all recordings, regardless of their label, using shift-invariant k-means; the continuous EEG is tokenized into count and transition features; and a downstream model is fit on those features.

\subsection{Shift-invariant k-means dictionary learning}
The single-channel dictionary learner follows Cano Achuri et al.~\cite{b_bow_mouse}, so we state it briefly; our additions are the extension to regional and cross-channel spatial atoms, and an exploration of the atom duration $D$ and dictionary size $K$. After bandpass filtering, the recording is split into non-overlapping windows of length $2D$ seconds, where $D$ is the atom duration, which lets an atom match at different phases within the window. Let $W_\tau$ extract a $P$-length sub-window starting at index $\tau$ from a length-$L$ window, and let its adjoint $S_\tau$ zero-pad a $P$-length atom and place it at offset $\tau$. The dictionary is a set of $K$ atoms $C=[c_1,\dots,c_K]$ of length $P<L$. We find the dictionary that best approximates each window with a single shifted and scaled atom,
\begin{equation}
\min_{C}\ \frac{1}{M}\sum_{i=1}^{M}\ \min_{\alpha,k,\tau}\ \big\|\tilde{x}_i - \alpha\, S_\tau(c_k)\big\|_2^2 ,
\label{eq:objective}
\end{equation}
where $M$ is the number of training windows. Because each window maps to one atom, this is a clustering problem (soft assignment gave worse results). For a fixed atom and shift the optimal non-negative scale has a closed form, and minimizing the residual reduces to maximizing a normalized inner product, giving the assignment
\begin{equation}
(k_i,\tau_i) = \arg\max_{k,\tau}\ \mathrm{sim}\big(W_\tau(\tilde{x}_i),\, c_k\big),
\label{eq:assign}
\end{equation}
where $\mathrm{sim}(a,b)=\langle a,b\rangle/(\|a\|_2\,\|b\|_2)$ is the cosine similarity. Cosine similarity makes the match amplitude-invariant, selecting an atom on shape alone, and the search over $\tau$ uses the fast Fourier transform. Each atom is updated as the mean of its shift-aligned windows, as in k-means. Each iteration costs $O(MKL)$; atoms start from random windows, and iteration stops when the mean squared atom change falls below $10^{-4}$ of the training-window variance, usually within 15 to 30 iterations.

\subsection{Preprocessing and gain-stripping}
Preprocessing decides what an atom encodes. We compared a per-channel z-score within each window plus a common average reference (CAR) against bandpass filtering only. When amplitude is kept in the match, two waveforms of the same shape but different amplitude split across several amplitude-specialized atoms. Removing amplitude, so matching is on shape, works better, since in EEG amplitude varies for non-physiological reasons such as electrode impedance, reference choice, and inter-subject scale. We call this gain-stripping. For multivariate atoms, CAR plus z-score is needed: without it, performance was worse (control vs AD F1 drops from 0.86 to 0.58), as the atoms learn amplitude artifacts rather than cross-channel shape.

\subsection{Feature extraction: counts and transitions}
The unigram counts follow the base method; the atom-to-atom transitions below are our second extension. After the dictionary is learned, every window in a segment is assigned to its best atom, which converts the segment into a single stream of atom tokens, one token per non-overlapping window in time order. We use hard assignment, with each window mapped to its single best atom, because a soft, similarity-weighted variant gave worse results. The simplest feature is the unigram, the count of each atom. To capture order we add bigrams and trigrams: an $n$-gram is a run of $n$ consecutive tokens in this stream, taken with a stride of one and encoded as a base-$K$ index. Counts are summed over a subject's sessions, and only $n$-grams that occur are kept, so the vector stays sparse. Counts are then weighted by term frequency and inverse document frequency (TF-IDF)~\cite{b_tfidf}, which up-weights atoms that are rare across segments, and the weighted vector is normalized to unit L2 norm.

The transition space grows as $K^n$ for $n$-grams, multiplied by the number of scales and dictionaries. To keep this manageable, we apply a variance-based selection that keeps the top few hundred features, 200 for the mouse and dementia tasks and 300 for TUEV, then L2-normalize. A minimum-frequency filter gave similar results, so we kept the simpler selection. In the multiscale variant we train separate dictionaries at more than one atom length (for example 0.25, 0.5, and 1.0 s), because rhythms live at different timescales, and concatenate their count vectors, using a smaller $K$ per scale to control the total dimension.

\subsection{Downstream models}
For the mouse genotype task the downstream model is k-means with two clusters. For the supervised tasks, the event and dementia tasks, the downstream model is a random forest~\cite{b_rf} with balanced class weights: 100 trees at depth 5 for the dementia task, and 300 to 400 trees for the event task, with balanced subsampling and square-root feature subsampling. On these supervised tasks we also tested an L2-regularized logistic regression and an RBF support vector machine~\cite{b_esl}; the random forest was the most consistent for Cohen kappa, while the linear models were sometimes better for balanced accuracy. The only learned quantities are the dictionary atoms and a small downstream model, so the parameter count is several orders of magnitude below the deep and foundation-model peers, which need millions to hundreds of millions of parameters and lose interpretability. The per-window assignment, n-gram counting, and weighting are independent across windows, channels, segments, and recordings, so the method parallelizes easily and runs on CPU.

\subsection{Evaluation protocols}
For the mouse task we cluster all data and score the adjusted Rand index (ARI) against genotype, with robustness checked over three seeds, twelve random hold-out splits, and a permutation null over 200 label shuffles ($p<0.01$). For the dementia task we use five-fold stratified cross-validation at the subject level, with subject-disjoint folds, and report accuracy and macro F1. For TUEV we use the official patient-disjoint split and report Cohen kappa, weighted F1, and balanced accuracy.

\section{Datasets}
\textbf{Mouse \textit{Scn8a} EEG.} Heterozygous \textit{Scn8a} mice (strain B6(C3Fe)-\textit{Scn8a}\textsuperscript{8J}/Frk, Jackson Laboratory stock 012945) exhibit spike-wave discharges characteristic of absence seizures~\cite{b_scn8a}. Ten heterozygous mice and six wild-type littermates were head-fixed and recorded with single-channel EEG over the left somatosensory cortex for about one hour (29 sessions, 1 to 2 per mouse). The EEG was sampled at 2000 Hz, low-pass filtered (Butterworth, 4th order, cutoff 55 Hz), and downsampled to 200 Hz, with a 0.5 Hz high-pass added as a separate preprocessing step. For clustering we keep one session per mouse, giving sixteen recordings, and use the extra sessions only to check stability. The downstream step is unsupervised k-means with two clusters. This dataset serves as a test of the temporal transition features and as a stress test of the low-data regime. We do not train a deep baseline here, since with sixteen recordings deep models overfit and human-pretrained models do not transfer.

\textbf{Dementia (ds004504).} Eyes-closed resting-state EEG in the 10-20 system, band-passed and artifact-cleaned as provided in the official dataset~\cite{b_ds004504}: 88 participants, 36 with Alzheimer's disease (AD), 29 controls, and 23 with frontotemporal dementia (FTD). We evaluate control vs AD, control vs FTD, and the three-way task. This dataset is where we test the regional and spatial (multivariate) atoms together with the temporal transitions, since here the signal lives in cross-channel topography rather than in discrete events. Detecting these conditions from EEG is attractive because EEG is cheaper and more accessible than imaging, which suits it to early screening.

\textbf{TUEV.} The Temple University EEG Event corpus~\cite{b_obeid} uses the 10-20 system in a bipolar montage, resampled to 250 Hz. Each annotation defines a 5 s window, taken from 2 s before to 3 s after onset, labeled with one of six events: spike and slow wave (SPSW), generalized periodic discharges (GPED), periodic lateralized discharges (PLED)~\cite{b_pled}, eye movement (EYEM), artifact (ARTF), and background (BCKG). Flagging these events automatically would offload work that clinicians and epileptologists otherwise do by hand. We follow the official patient-disjoint split, with roughly 84{,}000 training and 29{,}000 evaluation windows. This is the annotation-level case, with per-segment labels, and it tests the method against state-of-the-art deep and foundation models in the high-data regime where they are strongest. These baselines span from-scratch networks~\cite{b_eegnet,b_conformer,b_sparcnet} and pretrained models~\cite{b_biot,b_labram}.

\section{Results}
Because many settings can be varied across the different variants of the method, we sweep six axes. The first is univariate against multivariate atoms. The second is the dictionary source: on TUEV, a label-specific dictionary (balanced), a fully unsupervised dictionary tiled across whole recordings (whole), their union, and per-region dictionaries; on the dementia task, region-averaged univariate atoms (regional) and multivariate atoms with CAR and z-score (spatial). The third is the transition order, from unigram to four-gram. The fourth is the dictionary size $K$, between 6 and 64. The fifth is the downstream model, classification against clustering. The sixth is the assignment rule, hard against soft. Every dataset runs through the same sweep, and for each we report the best configuration together with the range of scores seen.

\subsection{Mouse EEG: genotype clustering}
\begin{figure}[t]
\centerline{\includegraphics[width=\columnwidth]{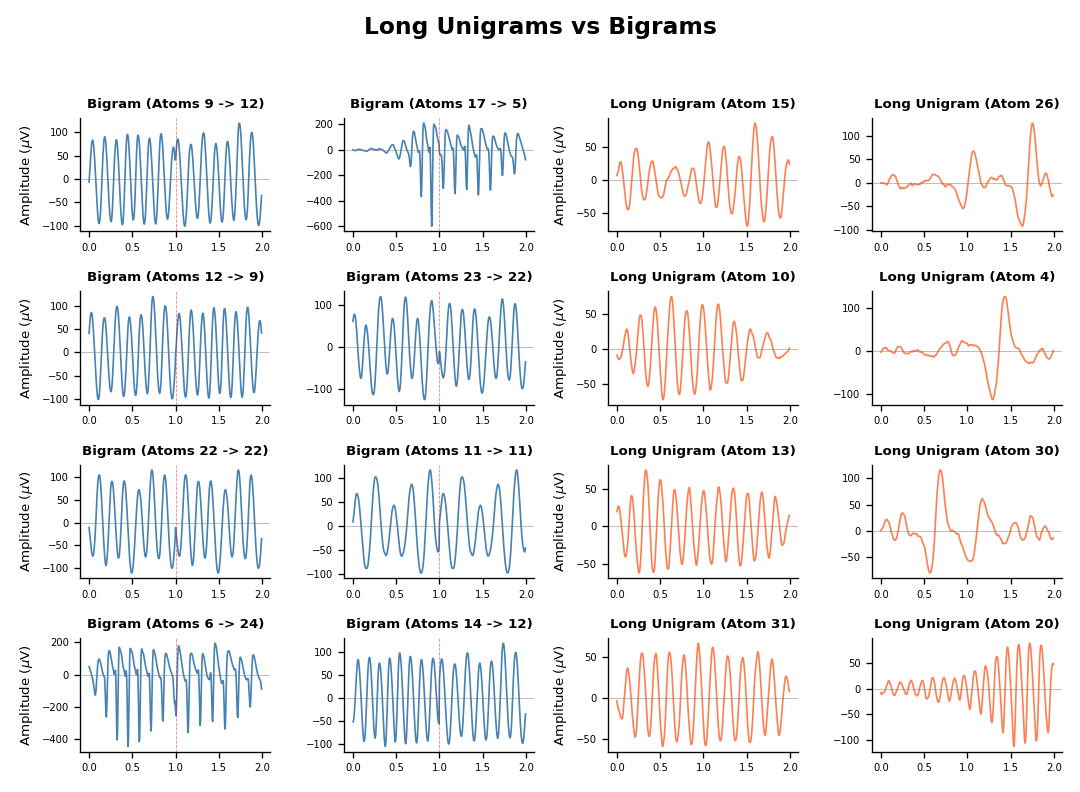}}
\caption{Bigram transitions (two 1 s atoms joined at the dashed line, left) against long unigram atoms (2 s, right). These are among the most discriminative atoms between the two genotypes rather than specific to one, and the paired short atoms appear sharper than the single long atom over the same span.}
\label{fig:mouse_morph}
\end{figure}

We start with the mouse dataset because its waveforms come out the sharpest, rather than smoothed into averaged prototypes as longer atoms are (Fig.~\ref{fig:mouse_morph}), and the two genotypes are the most distinct.

Unsupervised clustering on the atom features separates the genotypes perfectly, with an adjusted Rand index (ARI)~\cite{b_ari} of 1.0 (cluster sizes of six and ten, matching the wild-type and heterozygous split exactly). The best configuration uses a small dictionary of twelve 1.0 s atoms and unigram counts, reaching a silhouette~\cite{b_silhouette} of 0.78; six atoms still gives ARI 1.0 at silhouette 0.71, and the 64-atom baseline gives ARI 1.0 at silhouette 0.52. The result holds across three seeds and twelve hold-out splits (mean ARI 1.0), and the permutation null over 200 shuffles places the observed silhouette of 0.54 far from the null mean of 0.14, at $p<0.01$ and about 6.5 standard deviations out.

\begin{figure}[t]
\centerline{\includegraphics[width=\columnwidth]{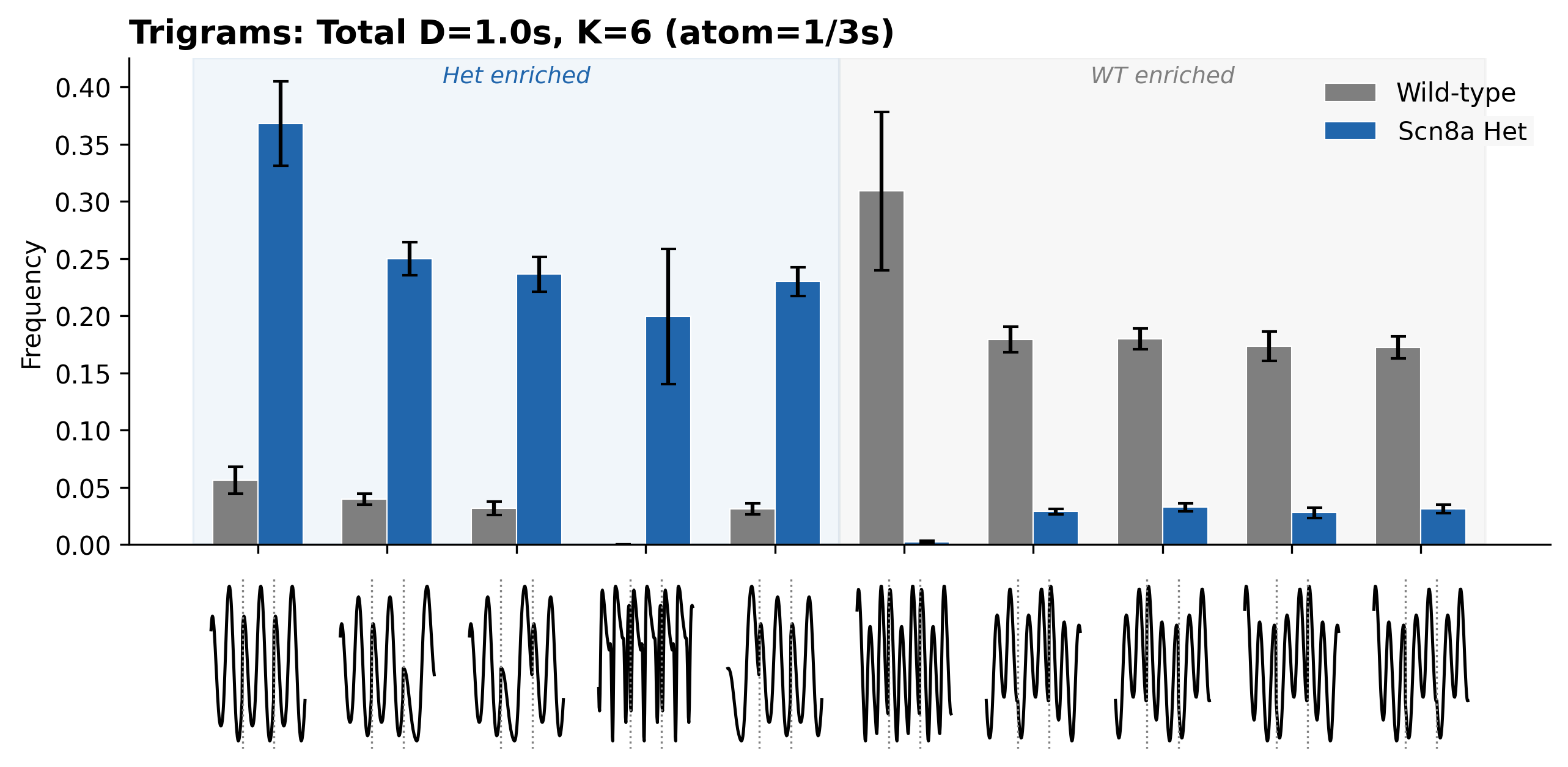}}
\caption{Per-genotype frequency of selected trigrams (three $1/3$ s atoms, $K=6$), wild-type against \textit{Scn8a} heterozygous, with the underlying waveform below each bar pair. Error bars show the spread across the model variants we tested.}
\label{fig:mouse_occ}
\end{figure}

The discriminative waveforms are interpretable (Figs.~\ref{fig:mouse_occ},~\ref{fig:mouse_morph}). Wild-type mice show faster background rhythms, while the \textit{Scn8a} heterozygous mice show slower rhythms together with spike and seizure-onset waveforms, consistent with the spike-wave phenotype of this line; in Fig.~\ref{fig:mouse_morph}, atom 17 captures a seizure onset and atoms 5, 6, and 24 capture spikes. Figure~\ref{fig:mouse_occ} shows that specific short wave sequences are several times more frequent in one genotype than the other, and the fourth heterozygous-enriched trigram is built from spike waveforms; this is what motivates modeling transitions, since the order in which waveforms recur differs by genotype.

Because ARI saturates at 1.0 across almost every configuration, the silhouette is a secondary read on compactness. A long atom is a poor substitute for a transition over the same span: at $K=64$ a 2 s unigram reaches silhouette 0.40, below a bigram of two 1 s atoms at 0.42; with a small dictionary, long unigrams fail outright, dropping to ARI 0.20 at 3.0 s and 0.75 at 4.0 s, while a bigram over the same span restores ARI 1.0 at silhouette 0.68. Combining two short waveforms captures multi-step structure that one stretched waveform blurs away (Fig.~\ref{fig:mouse_morph}). The transition also reads more cleanly, an interpretability gain on top of the accuracy one. Both benefits depend on a small dictionary: at $K\ge 128$ the transition space explodes and bigram ARI collapses to 0.08. Sparse k-means and its variants mitigate this but do not fully solve it, so the feature selection would need more care, which is outside our scope.

The method also needs very little data. A minimum-duration sweep gives ARI 1.0 at 120 s, about 0.90 at 60 s, and around 0.60 at 20 to 30 s, falling to chance only at 10 s. One minute of single-channel EEG already gives a near-perfect split, and two minutes makes it exact.

\subsection{Dementia classification}
We now move from one channel to multi-channel montages, and to a task with no single discrete biomarker, where the signal lives in cross-channel shape and its changes over time.

Control vs AD reaches a macro F1 of 0.86 and a balanced accuracy of 86.2 percent (chance is 50 percent for the binary tasks and 33 percent for the three-way), from multivariate atoms trained with CAR and z-score, 0.5 s atoms combined into bigrams, and a small dictionary of eight atoms. Control vs FTD reaches F1 0.79 (balanced accuracy 80.9 percent). The three-way task, AD vs FTD vs control, is much harder, and both scores are lower at F1 0.55 and balanced accuracy 62.3 percent, but on par with other state-of-the-art models. Across the full sweep, control vs AD ranged from F1 0.75 to 0.86, control vs FTD from 0.63 to 0.79, and the three-way task from 0.45 to 0.55. Table~\ref{tab:ad}(a) compares these with prior work on ds004504, which uses deep or connectivity-based models under different validation protocols, so the comparison is only indicative; bag-of-waves lands in the same range without any deep learning or spectral features.

The spatial atoms slightly outperform the regional ones (Table~\ref{tab:ad}(b)), reaching F1 0.86 on control vs AD against 0.79 for regional. The gain comes from the cross-channel information the spatial atoms capture, a form of connectivity that has been shown to be informative for dementia~\cite{b_mlinaric}.

Unigrams alone give a strong baseline (F1 around 0.78 on control vs AD); bigrams give the best result at 0.86, so the transition between two short brain states is more telling than either state alone; trigrams and four-grams lead to feature explosion, since the feature space grows quickly and the top-200 selection can no longer find the signal. Short windows of 0.5 to 1.0 s and eight to sixteen atoms are best. The simplest configuration, univariate atoms with unigram counts, is the closest match to the base method~\cite{b_bow_mouse}, which used single-channel unigrams without transitions, and stays just below F1 0.80, while multivariate spatial atoms with bigram transitions reach 0.86, so both additions are needed.

\begin{figure}[t]
\centerline{\includegraphics[width=\columnwidth]{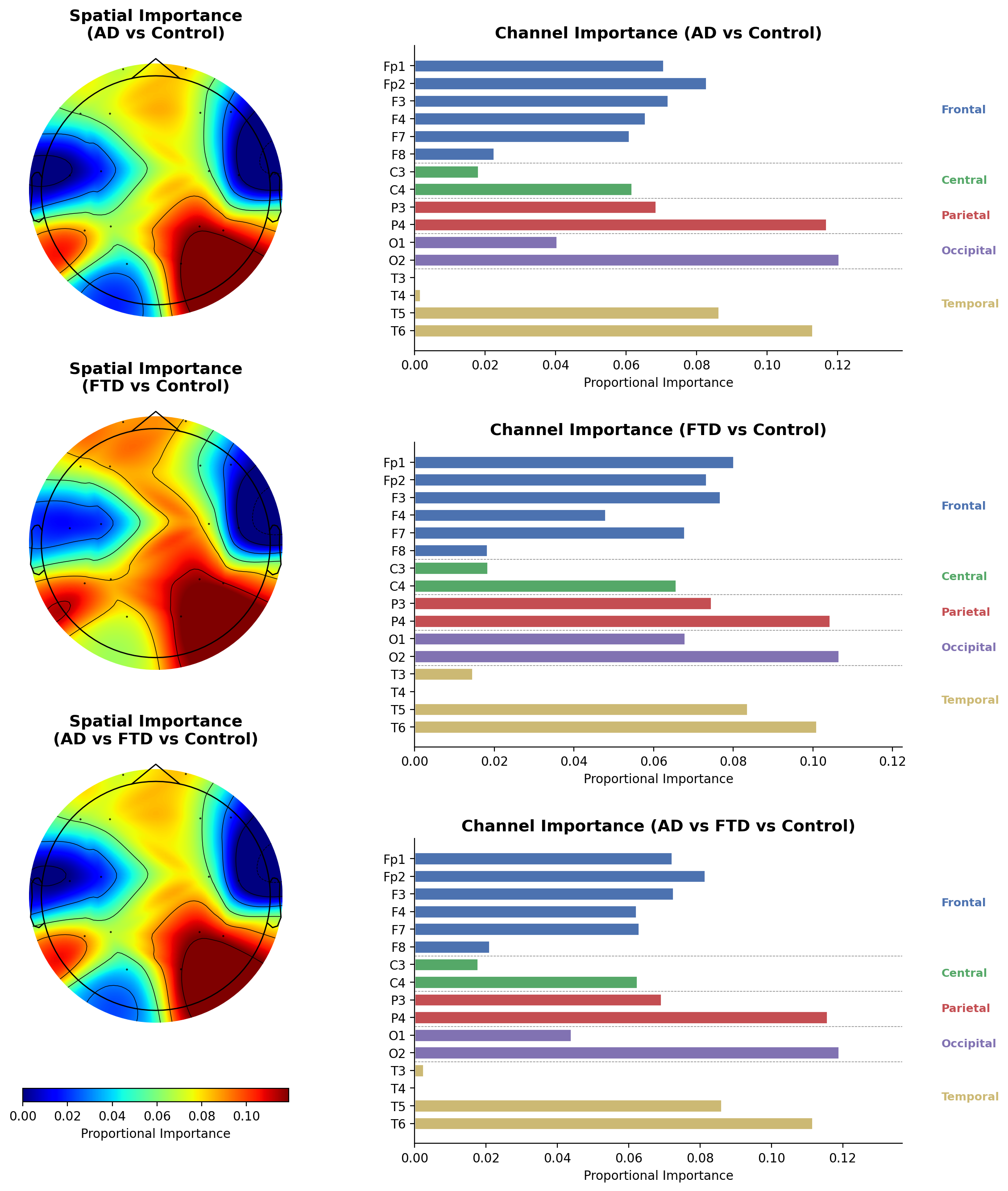}}
\caption{Per-channel importance for the spatial pipeline, as scalp maps (left) and grouped bars (right), for the three tasks.}
\label{fig:ad_importance}
\end{figure}

\begin{figure}[t]
\centerline{\includegraphics[width=\columnwidth]{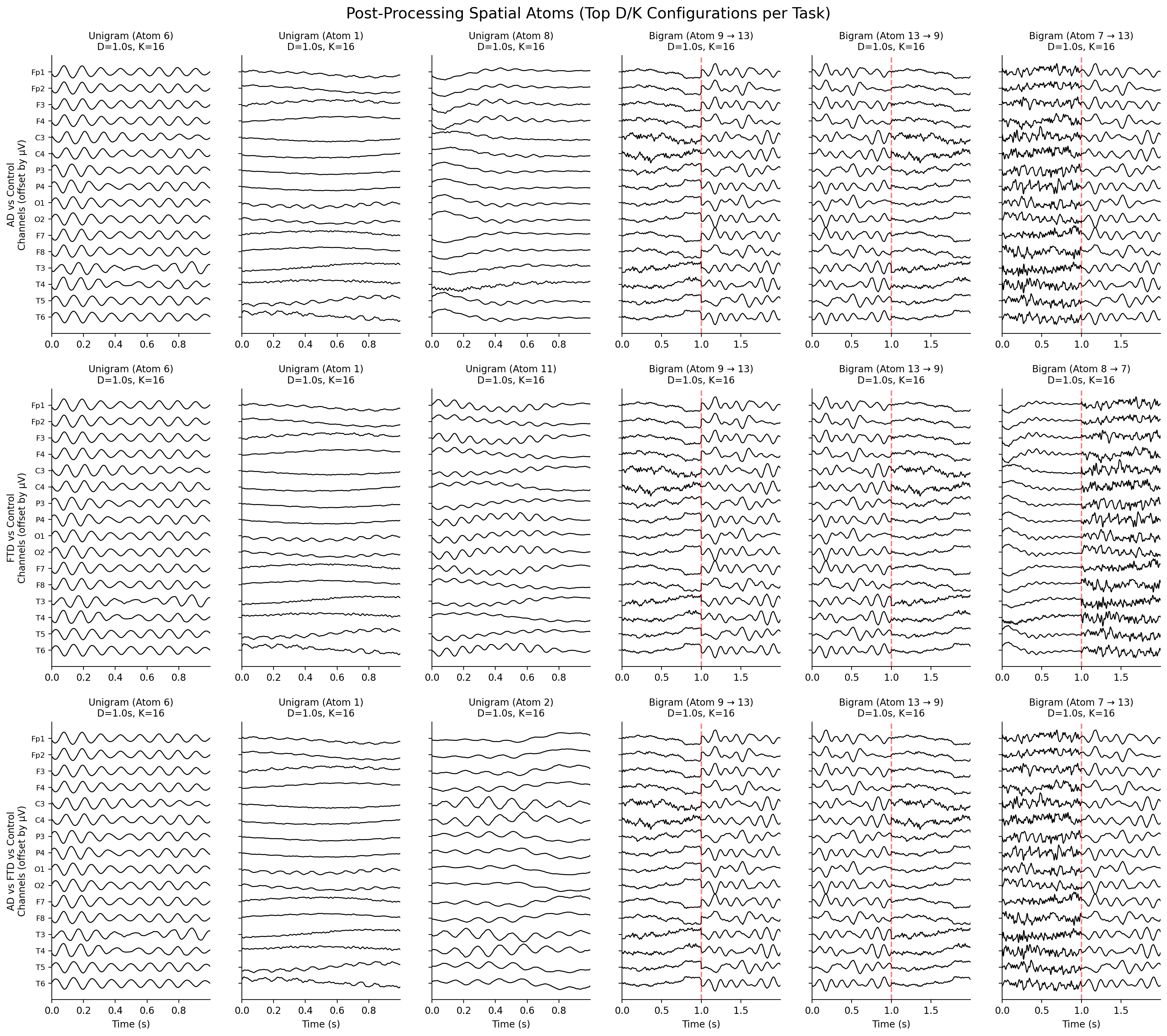}}
\caption{Multivariate spatial atoms after CAR and z-score, drawn as multichannel traces ($D=1.0$ s, $K=16$). The first three columns are unigram atoms; the last three are bigram transitions, with the dashed line marking the atom boundary.}
\label{fig:ad_topo}
\end{figure}

The interpretable signal here is both spatial and temporal. Channel importance (Fig.~\ref{fig:ad_importance}), from the random forest's Gini importance or the linear model's coefficients, concentrates over posterior channels (occipital O2, parietal P4) and temporal channels (T5, T6), with a frontal contribution, matching the known resting-state signature of Alzheimer's disease and frontotemporal dementia~\cite{b_qeeg,b_alpha}. The atoms themselves (Fig.~\ref{fig:ad_topo}) confirm that with CAR and z-score the spatial atoms encode differential cross-channel topography and its change over time, not single-channel amplitude. Independently trained univariate atoms per region show similar temporal patterns.

Grouping the learned atoms by dominant frequency and counting per patient, the classic alpha slowing of Alzheimer's disease~\cite{b_qeeg,b_alpha} appears as a change in which waveforms recur. Fast alpha atoms (10.5 to 11.5 Hz) occur about 1.5 times more often in controls than in the AD group, while slow alpha and theta atoms (7 to 9 Hz) occur slightly more often in the AD group. The method recovers the well-known spectral signature, but by counting morphological tokens a reader can plot and inspect rather than a band-power ratio. The temporal reading is that short, sub-second atoms carry the signal, which fits the bigram result that the information is in fast transitions between short states. A traditional-features baseline combining band power, connectivity, Hjorth parameters, and graph summaries reached only 76.9 percent balanced accuracy on control vs AD under the same subject-level cross-validation.

The pipeline is robust to scarce data. Control vs AD F1 holds near 0.75 using only the first tenth of each recording, and with only half the patients (about fourteen per class), degrading to chance only below about five per class, where the supervised step, not the atoms, runs out of examples. Learning a separate dictionary per diagnostic label and concatenating them (Table~\ref{tab:ad}(c)) helps in some settings but hurts the best single-scale result, dropping control vs AD from 0.86 to 0.76, and some gains are matched by enlarging the global dictionary. We therefore keep the global, label-free dictionary throughout, which also generalizes to recordings whose diagnosis is unknown.

\begin{table}[t]
\caption{Alzheimer's disease (ds004504) results. (a) Balanced accuracy against prior work; validation differs (LOSO, leave-one-subject-out; subj., subject-based cross-validation; ours, five-fold subject-level CV), so the comparison is only indicative. (b) Best macro F1 per pipeline; spatial is multivariate atoms with CAR and z-score. (c) Global label-free vs.\ label-refined dictionary, macro F1.}
\label{tab:ad}
\centering
\footnotesize
\setlength{\tabcolsep}{4pt}
\textit{(a) Balanced accuracy vs.\ prior work}\\[2pt]
\begin{tabular}{llccc}
\toprule
\textbf{Method} & \textbf{Val.} & \textbf{C vs AD} & \textbf{C vs FTD} & \textbf{3-class} \\
\midrule
DICE-Net~\cite{b_dicenet} & LOSO & 83.3 & 75.0 & -- \\
FC ensemble~\cite{b_mlinaric} & LOSO & 73.9 & 71.2 & -- \\
FC best single~\cite{b_mlinaric} & LOSO & 84.6 & 82.7 & -- \\
Deep learning~\cite{b_delpup} & subj. & -- & -- & 58.1 \\
Traditional features (ours) & 5-fold & 76.9 & 69.2 & 55.7 \\
\textbf{Bag-of-waves (ours)} & 5-fold & \textbf{86.2} & \textbf{80.9} & \textbf{62.3} \\
\bottomrule
\end{tabular}\\[5pt]
\textit{(b) Best macro F1 per pipeline}\\[2pt]
\begin{tabular}{lcc}
\toprule
\textbf{Task} & \textbf{Regional} & \textbf{Spatial} \\
\midrule
control vs AD & 0.79 & \textbf{0.86} \\
control vs FTD & 0.77 & \textbf{0.79} \\
AD vs FTD vs control & 0.51 & \textbf{0.55} \\
\bottomrule
\end{tabular}\\[5pt]
\textit{(c) Global vs.\ label-refined dictionary (macro F1)}\\[2pt]
\begin{tabular}{lcc}
\toprule
\textbf{Task (setting)} & \textbf{Global} & \textbf{Label-refined} \\
\midrule
AD vs control (single-scale) & 0.86 & 0.76 \\
FTD vs control (single-scale) & 0.74 & 0.76 \\
AD vs control (multiscale) & 0.72 & 0.78 \\
AD/FTD/control (multiscale) & 0.47 & 0.53 \\
\bottomrule
\end{tabular}
\end{table}

\subsection{TUEV: event classification}
TUEV has dense per-segment annotations for six event types, so it tests whether the same tokens and transitions handle annotation-level tasks, not only whole-recording labels. Because the labels are expert event types, it also lets us check against ground truth that the label-free atoms recover clinical morphologies (the periodic discharges PLED and GPED).

The best result comes from the union of the balanced and whole dictionaries, both gain-stripped, with unigram, bigram, trigram, and multiscale counts fed to a random forest. It reaches a Cohen kappa~\cite{b_kappa} of 0.44, a weighted F1 of 0.72, and a balanced accuracy of 0.44. Across the full sweep, kappa ranged from 0.36 to 0.44.

The two dictionaries are complementary (Table~\ref{tab:tuev}(a)), which is why their union wins. The balanced dictionary is strong on GPED (F1 around 0.62) but weak on PLED (around 0.33); the whole dictionary is the opposite, both near 0.46; the union inherits both (GPED 0.57, PLED 0.41). Hard assignment matters most here: an earlier soft variant reached only about 0.23 PLED F1. Transitions help: for the union, adding higher-order features raises kappa from about 0.43 to 0.44, and the gain is strongest only when the two vocabularies are combined. The balanced dictionary at $K=16$ marginally beats $K=32$, though we default to $K=32$ for vocabulary richness.

\begin{figure*}[t]
\centerline{\includegraphics[width=\textwidth]{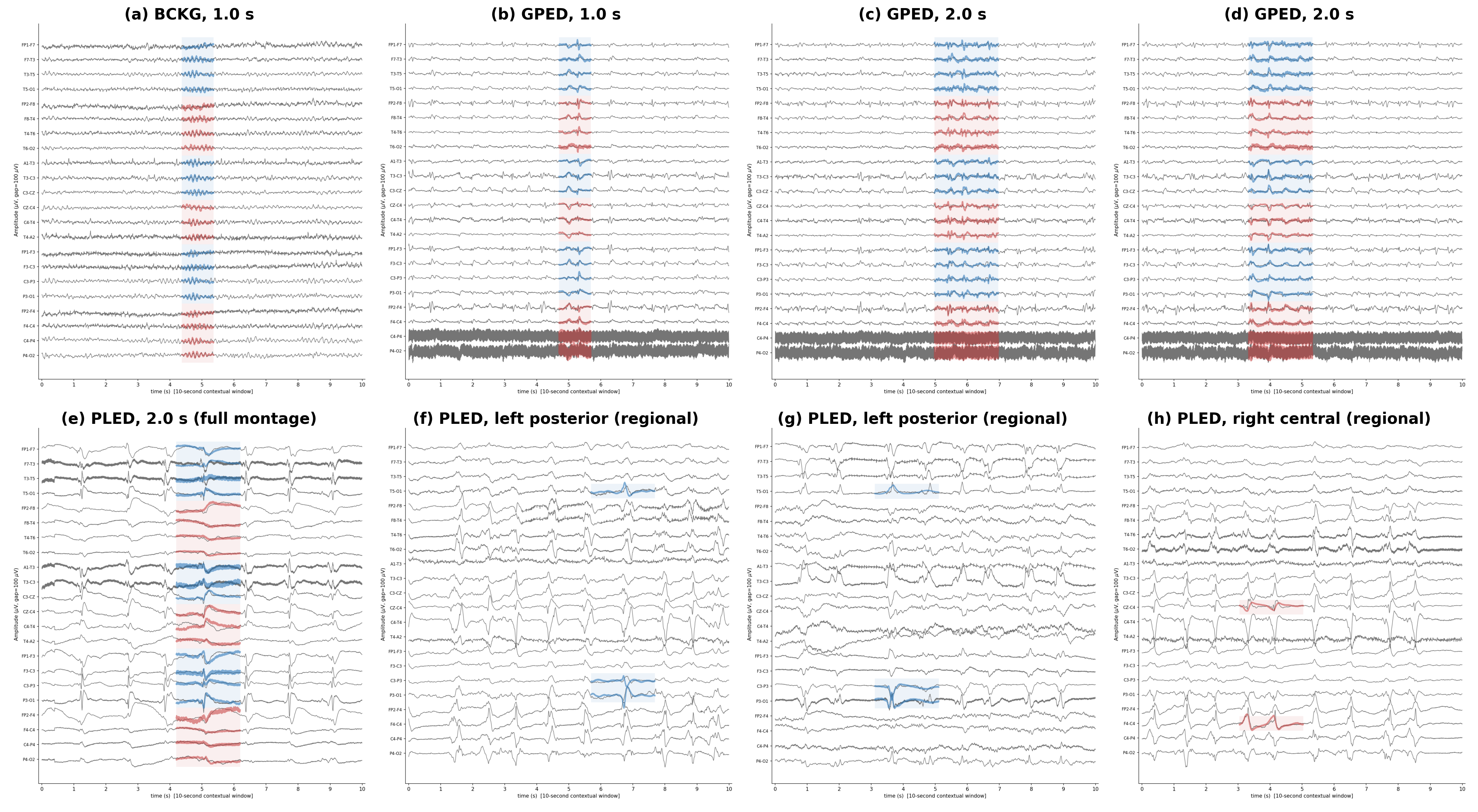}}
\caption{Label-free atoms recovered on TUEV, each overlaid as a thick colored trace on an example recording within a 10-second context. Blue marks the left-hemisphere bipolar chains, red the right. Top row: a rhythmic background atom (a) and generalized periodic discharge atoms at 1.0 s and 2.0 s (b to d). Bottom row: periodic lateralized discharge atoms, from the full montage (e) and localized to one lobe and side from the per-region dictionaries (f to h).}
\label{fig:tuev_atoms}
\end{figure*}

Fig.~\ref{fig:tuev_atoms} shows the highest-importance atoms recovering the annotated morphologies: a rhythmic bilateral background atom, bilateral synchronous GPED complexes captured at both the 1.0 s and 2.0 s scales, and PLED atoms that localize to one lobe and side using the per-region dictionaries. We see that the label-free dictionary recovers clinically defined events, not just counts that separate the classes. We do not show a spike and slow wave (SPSW) panel: few patients contributed SPSW events, so those atoms are not reliable and the class is poorly solved by any variant. A single spike is also very brief, under about 70 ms and shorter than the atom lengths we swept, so shorter atoms might capture it better; we leave this to future work.

Table~\ref{tab:tuev}(b) places our result among published deep-learning and foundation models. Our kappa of 0.44 is competitive and sits above FFCL and the transformer baselines and just below vanilla BIOT~\cite{b_biot}. The higher entries, such as the pretrained LaBraM~\cite{b_labram} and other foundation models, exploit a large labeled corpus with millions of parameters; our method uses far fewer and is built for the small-data settings where those models are not even feasible to train. These models also act on short fixed windows, so a long recording must be split and pooled, whereas bag-of-waves summarizes a whole recording by its waveform rates and transitions, suiting recording-level questions like the mouse genotype and dementia diagnosis.

\begin{table}[t]
\caption{TUEV six-class results. (a) Best configuration per dictionary source ($K=32$, gain-stripped; U/B/T/M denote unigram/bigram/trigram/multiscale). (b) Comparison with published models, ascending by Cohen's $\kappa$; literature values are mean $\pm$ SD, ours are point estimates.}
\label{tab:tuev}
\centering
\footnotesize
\setlength{\tabcolsep}{4pt}
\textit{(a) Dictionary source}\\[2pt]
\begin{tabular}{lcccc}
\toprule
\textbf{Source} & \textbf{Features} & \textbf{$\kappa$} & \textbf{wF1} & \textbf{bal-acc} \\
\midrule
balanced & U+B+T+M & 0.43 & 0.71 & 0.41 \\
whole & U & 0.40 & 0.72 & 0.42 \\
\textbf{union} & \textbf{U+B+T+M} & \textbf{0.44} & \textbf{0.72} & \textbf{0.44} \\
\bottomrule
\end{tabular}\\[5pt]
\textit{(b) Comparison with published models}\\[2pt]
\scriptsize
\begin{tabular}{lccc}
\toprule
\textbf{Method} & \textbf{Bal. Acc.} & \textbf{Cohen's $\kappa$} & \textbf{Weighted F1} \\
\midrule
FFCL~\cite{b_ffcl} & $0.40 \pm 0.01$ & $0.37 \pm 0.02$ & $0.68 \pm 0.01$ \\
ST-Transformer~\cite{b_sttransformer} & $0.40 \pm 0.02$ & $0.38 \pm 0.03$ & $0.68 \pm 0.02$ \\
CNN-Transformer~\cite{b_cnntransformer} & $0.41 \pm 0.02$ & $0.38 \pm 0.01$ & $0.69 \pm 0.03$ \\
ContraWR~\cite{b_contrawr} & $0.44 \pm 0.03$ & $0.39 \pm 0.02$ & $0.69 \pm 0.01$ \\
SPaRCNet~\cite{b_sparcnet} & $0.42 \pm 0.03$ & $0.42 \pm 0.02$ & $0.70 \pm 0.01$ \\
\textbf{Bag-of-waves (ours)} & $\mathbf{0.44}$ & $\mathbf{0.44}$ & $\mathbf{0.72}$ \\
Vanilla BIOT~\cite{b_biot} & $0.47 \pm 0.01$ & $0.45 \pm 0.03$ & $0.71 \pm 0.02$ \\
BIOT~\cite{b_biot} & $0.47 \pm 0.04$ & $0.49 \pm 0.04$ & $0.74 \pm 0.02$ \\
LaBraM-Base~\cite{b_labram} & $0.47 \pm 0.09$ & $0.51 \pm 0.04$ & $0.75 \pm 0.02$ \\
TFM-Tokenizer~\cite{b_tfm} & $0.49 \pm 0.05$ & $0.53 \pm 0.03$ & $0.76 \pm 0.02$ \\
SEBSFormer~\cite{b_sebsformer} & $0.66 \pm 0.01$ & $0.69 \pm 0.01$ & $0.84 \pm 0.01$ \\
\bottomrule
\end{tabular}
\end{table}

\subsection{Linear classifiers and bag-of-waves against hidden Markov models}
The careful reader may have noticed that, although the focus is on interpretability, the supervised results use a random forest, which is less interpretable than a linear model. Since the features are already discrete, named tokens, we also fit a logistic regression, which gives each atom and transition a signed per-class coefficient that can be read directly. This costs little (Table~\ref{tab:alt}(a)): it stays within a few points of the random forest on the binary dementia tasks, is higher on the three-way task, and matches it on TUEV balanced accuracy while trading some kappa, so a fully interpretable classifier is a knob, not a constraint. A reviewer might also expect a comparison with hidden Markov models, since the transitions are a discrete counterpart to microstate and HMM analyses. We ran a generative per-class categorical HMM and an unsupervised HMM whose Viterbi-decoded~\cite{b_viterbi} transition matrix and state-occupancy vector feed a random forest~\cite{b_hmm_baum}, sweeping 2 to 16 hidden states. Both underperform the direct empirical counts (Table~\ref{tab:alt}(b)): forcing the sequence through a few latent states blurs the specific transitions, the first-order Markov assumption discards the longer-range periodicity the trigrams keep, and the generative HMM is fit to model the sequence rather than to separate the classes.

\begin{table}[t]
\caption{Alternative downstream models on the same token features. (a) Interpretable logistic regression (LR), multinomial on the three-way task, vs.\ random forest (RF); dementia rows use the best AD configuration ($D{=}0.5$ s, bigram, $K{=}8$, spatial), TUEV rows use the union dictionary. (b) Hidden Markov model baselines, the better of a generative per-class model and a feature extractor feeding a random forest, vs.\ bag-of-waves.}
\label{tab:alt}
\centering
\footnotesize
\setlength{\tabcolsep}{5pt}
\textit{(a) Logistic regression vs.\ random forest}\\[2pt]
\begin{tabular}{llcc}
\toprule
\textbf{Task} & \textbf{Metric} & \textbf{RF} & \textbf{LR} \\
\midrule
AD vs control & macro F1 & 0.86 & 0.81 \\
FTD vs control & macro F1 & 0.79 & 0.77 \\
AD vs FTD vs control & macro F1 & 0.55 & 0.57 \\
TUEV (6-class) & Bal.\ acc & 0.44 & 0.43 \\
TUEV (6-class) & Cohen's $\kappa$ & 0.44 & 0.31 \\
\bottomrule
\end{tabular}\\[5pt]
\textit{(b) Hidden Markov model vs.\ bag-of-waves}\\[2pt]
\begin{tabular}{lcc}
\toprule
\textbf{Task} & \textbf{Bag-of-waves} & \textbf{Best HMM} \\
\midrule
Dementia, control vs AD (F1) & 0.86 & 0.78 \\
TUEV, 6-class (Cohen's $\kappa$) & 0.44 & 0.25 \\
\bottomrule
\end{tabular}
\end{table}

Table~\ref{tab:synthesis} brings the three results together with the best configuration for each.

\begin{table}[htbp]
\caption{Cross-dataset summary: best configuration per benchmark.}
\label{tab:synthesis}
\centering
\footnotesize
\setlength{\tabcolsep}{3.5pt}
\begin{tabular}{llll}
\toprule
\textbf{Dataset} & \textbf{Downstream} & \textbf{Best config} & \textbf{Score} \\
\midrule
Mouse & k-means & $K{=}12$, 1.0\,s, unigram & ARI$\,{=}\,1.0$ \\
AD & random forest & spatial, $K{=}8$, bigram & F1$\,{=}\,0.86$ \\
TUEV & random forest & union, $K{=}32$ & $\kappa{=}0.44$ \\
\bottomrule
\end{tabular}
\end{table}

\section{Conclusion}
Bag-of-waves approaches the performance of state-of-the-art deep and foundation models across recording setups and clinical questions while staying low-data, lightweight, and interpretable, with no deep learning, pretraining, or predefined spectral features. Its two extensions, atom-to-atom transitions and multivariate spatial atoms, let one label-free representation handle settings as different as single-channel mouse genotyping, resting-state dementia, and clinical event annotation. Its value is the balance it strikes among accuracy, compute, interpretability, and data efficiency.

\subsection*{Limitations and future work}
Each atom is the mean of many shift-aligned windows, so it is a smoothed prototype. On stereotyped events like the mouse spikes the atoms stay crisp, but on routine clinical and resting-state data residual artifact and variability blur the average, so the recovered waveform looks smoothed relative to the true biomarker. Classification still holds, but the fine morphology is partly washed out, and this trades against the small dictionary we otherwise recommend, since a smaller dictionary averages more windows per atom. A second limitation is inherited from the base method: each window is still explained by one shifted and scaled atom, so two events overlapping within a window are not separated, and the cosine-similarity assignment gives convenient prototypes rather than an optimal sparse code. The method also depends on recurrence: rare events and very short recordings give unstable tokens, which is why the SPSW class is not solved and why the mouse task fails below about one minute.

We do not test cross-dataset transfer, and we choose hyperparameters per task; both are future work. The vocabulary is also discrete, with no notion of similarity between atoms; future work could embed the atoms in a continuous space learned from their co-occurrence and transition statistics, keeping interpretability while allowing soft assignment across related atoms and possibly better transfer. The feature selection is a simple variance filter, kept simple to isolate the contribution of the dictionary. Finally, while the method recovers known morphologies and surfaces one exploratory observation, it does not establish a new clinically validated biomarker; the natural next step is biomarker discovery and disease subtyping on other datasets, with prospective and external validation, which is where an interpretable, label-free, low-data method should be most useful.

\end{document}